\documentclass[runningheads]{llncs}
\usepackage[T1]{fontenc}
\usepackage{graphicx}
\usepackage{mathptmx}
\DeclareMathAlphabet{\mathcal}{OMS}{cmsy}{m}{n}
\usepackage{soul}\setuldepth{article}
\def\hb{\hbox to 11.5 cm{}}

   \usepackage[neveradjust]{paralist}
  \usepackage{soul, color}
  \sethlcolor{green}
\newtheorem{defs}{\textbf{Definition}}

\usepackage{tcolorbox}

\begin{document}
\title{Discerning and Characterising Types of Competency Questions for Ontologies}
%
%

\author{C. Maria Keet\inst{1}\orcidID{0000-0002-8281-0853} \and
Zubeida Casmod Khan\inst{2}\orcidID{0000-0002-1081-9322}}
\authorrunning{Keet and Khan}
%
\institute{Department of Computer Science, University of Cape Town, South Africa
\email{mkeet@cs.uct.ac.za}\\
 \and
Council for Scientific and Industrial Research, Pretoria, South Africa\\
\email{zdawood@csir.co.za}}
\maketitle              
\begin{abstract}
Competency Questions (CQs) are widely used in ontology development by guiding, among others, the scoping and validation stages. 
However, very limited guidance exists for formulating CQs and assessing whether they are good CQs, 
leading to issues such as ambiguity and unusable formulations. To solve this, one requires insight into the nature of CQs for ontologies and their constituent parts, as well as which ones are not. We aim to contribute to such theoretical foundations in this paper, which is informed by analysing questions, their uses, and the myriad of ontology development tasks.
This resulted in a first  Model for Competency Questions, which comprises five main types of CQs, each with a different purpose: Scoping (SCQ), Validating (VCQ), Foundational (FCQ), Relationship (RCQ), and Metaproperty (MpCQ) questions. 
This model enhances the clarity of CQs and therewith aims to improve on the effectiveness of CQs in ontology development, thanks to their respective identifiable distinct constituent elements. We illustrate and
evaluate them with a user story and demonstrate where which type can be used in ontology development tasks. To foster use and research, we created an annotated repository of 438 CQs, the Repository of Ontology Competency QuestionS (ROCQS), incorporating  an existing CQ dataset and new CQs and CQ templates, which further demonstrate distinctions among types of CQs.

\end{abstract}
\section{Introduction}
\label{sec:intro}

Competency questions (CQs) are broadly used within a variety of domains such as education to  design curricula using Bloom's Taxonomy, performance evaluation to assess an employee's performance, or assessing fitness of interrogation in a trial  \cite{Lyon01,Williams96}. Within ontology development, CQs are used throughout the processes to guide the development, such as in the NeOn methodology \cite{Suarez08} and test-driven development \cite{KL16}, including scoping the ontology \cite{Uschold96}, 
aligning ontologies \cite{Thieblin18}, validating content coverage \cite{Bezerra17,Bezerra13,KL16}, and to assist in interrogating the ontological nature of an entity when aligning a domain entity to an entity in a foundational ontology \cite{BKKM23}. 

However, CQs are used in different ways and sometimes incorrectly and ambiguously \cite{AK23,WISNIEWSKI2019100534},  and ontology engineers have difficulties working with CQs in a systematic way from the beginning to the end of the ontology development task  
 \cite{10.1007/978-3-031-47262-6_3}. In some cases CQs are not answerable due to too imprecise wording for it to be amenable to a formalisation into a logic or query language \cite{PWLK20} or due to expressiveness limitations or content coverage limitations. Further, existing ontology development methodologies make little use of elicitation and modelling techniques for CQs. Consequently, ontology developers have limited resources for developing CQs,  other than the CLaRO controlled natural language and tool for CQ authoring \cite{KMA19} and preliminary automation of CQ generation \cite{AK23,Alharbi2023AnEI}. Incomplete or ambiguous CQs, in turn, may result in incoherent or incomplete ontologies and the ontology-based systems that use them. Despite recent increase in popularity in adoption of CQs, what they exactly are, or should be, or what `exemplary' or `good' CQs should look like, remains unclear, let alone the idea that there may be different types of CQs. 

We aim to contribute to theoretical insights into CQs for ontologies that in turn may 
address these problems by devising a first categorisation of CQs into five different principal types of CQs, being Scoping (SCQ), Validating (VCQ), Foundational (FCQ), Relationship (RCQ), and Metaproperty (MpCQ) questions. They each have a distinct purpose and, consequently, different characteristic constituents of a question of that type, showing that they are indeed distinct and distinguishable. 
This is further made clear in the model and the formalisation of each type of CQ. In consulting the literature and ontologies to  attempt to source examples, it showed a dearth of foundational, relationship, and metaproperty CQs, and therefore we  created more CQs of those types, which further demonstrate their distinctions for the various ontology development tasks. To foster use, reuse, and further research, we 
(1) describe where the different types of CQs would typically be used at different stages and tasks in the NeOn methodology, (2) demonstrate use with a use case, and (3) developed a FAIR-compliant repository of CQs, which categorises the CQs by type and their key constituents, incorporating both an existing CQ dataset and new CQs and CQ templates.  

The remainder of the paper is structured as follows. Section~\ref{sec:relwork} discusses related work. This is followed by the introduction of the Questions for Ontologies (QuO) model  in Section~\ref{sec:model}. Use-cases, their use in an ontology development methodology, and the repository of CQs, ROCQS, are elaborated on in Section~\ref{sec:use}. Lastly, we discuss in Section~\ref{sec:disc} and conclude in Section~\ref{sec:concl}.

\section{Related Work}
\label{sec:relwork}

The first mention of CQs for ontologies can be traced back to  1996, as ontologies and ontology engineering was emerging as a field. Uschold and Gruninger were the first to describe CQs as those that ``specify the requirements for an ontology and as such are the mechanism for characterising the ontology design search space'' \cite{Uschold96}. 
They distinguished between informal and formal CQs. Informal CQs must be consistent with the axioms in the ontology and serve as constraints on what the ontology can be, they assert, rather than determining a particular design with its corresponding ontological commitments. 

Wisniewski et al. \cite{WISNIEWSKI2019100534} collected and analysed 234 informal CQs for several ontologies in different subject domains, analysed those, and translated them in SPARQL-OWL where possible, having created a first dataset of CQs, their patterns, and SPARQL-OWL queries \cite{PWLK20}. The computational analysis of the data showed there to be no 1:1 mapping to OWL constructs as had previously been assumed, which is partly due to modelling styles and partly due to different natural language sentence constructions for the same formalisation.  
Those data-driven patterns also served as input to the development of the CLaRO controlled natural language to author CQs \cite{KMA19}. Since CLaRO is derived from existing CQs of domain ontologies, it is unclear whether they are also suitable for CQs such as those dealing with meta-properties in ontologies or CQs relevant for foundational ontologies. This also remains the question with their more recent work on automating CQ generation, since it has CLaRO in the loop as part of the CQ generation pipeline \cite{AK23}.  
A LLM-based method to generate CQs for existing ontologies to foster ontology reuse showed promising results, but also that some CQs are incorrect, could not be validated by the developers, or were not matched or usable due to issues such as ambiguity and subjectivity \cite{Alharbi2023AnEI}, and similarly for explorations of prompting LLMs for CQs for a new ontology \cite{Rebboud24}.  

Monfardini et al. \cite{10.1007/978-3-031-47262-6_3} surveyed the use of CQs in ontology engineering recently, alongside their benefits and challenges. The results show that CQs have been considered useful and have helped mainly to define an ontology's scope and to evaluate an ontology, and that, mostly, CQs were authored iteratively and refined along the ontology development process. The authors conclude that the lack of practical and detailed guidelines and supporting tools contributes to the difficulties faced by ontology engineers. This might be a cause of Guizzardi and Guarino's \cite{DBLP:journals/corr/abs-2304-11124} claim of an issue of vague or unspecific competency questions that hinders ontology development. They also contend that CQs typically serve as mere lookup queries within a model's structure but that their true value lies in their ability to guide ontology construction when framed as genuine context-dependent requests for explanation. This indicates CQs are expected to serve multiple purposes.  

To some extent, the latter emerges also from ontology engineering methodologies. Among others, Tropos has been proposed to visually capture and refine CQs as goals, enabling analysis through relationships with plans and resources \cite{Fernandes2011UsingGM}. Though neither Tropos, nor aforementioned CLaRO, consider validation with CQs. 
The NeOn methodology \cite{Suarez08} and test-driven development \cite{KL16} do assume CQs for both scoping and validation, as do Bezerra et al. for the latter \cite{Bezerra17,Bezerra13}. They also have been proposed for use in other ontology development tasks separately, such as to assist aligning ontologies \cite{Thieblin18,BKKM23}, and thus that CQs may be of use in various ways throughout the ontology engineering process.

In sum, there is some limited assistance in CQ authoring for ontologies, a notion of different tasks and goals, and some confusion, let alone a notion of quality of CQs, but that nonetheless they are already perceived to be of use. 

\section{Model for Competency Questions}
\label{sec:model}
\label{sec:cqdefs}

\subsection{Preliminary analysis: purposes and core elements}
\label{sec:prelim}

As a first step towards model development, we elicitate possible purposes of CQs in relation to ontologies and entities represented in them, and elements that are relevant to CQs.

\subsubsection{Purposes of CQs}

Let us proceed to be more specific about  direct and underlying motivations, which can also be referred to informally as {\em purpose}, {\em goal}, {\em function}, or {\em intent} beyond a mere `information elicitation' function of a question, and therewith working towards a {\em why} to ask and answer a CQ. Consider:
\begin{compactitem}[--]
\item {\em Demarcation of the subject domain} for the ontology; e.g., in ``Which plant parts does Pumba the warthog eat?'', it helps demarcating the subject domain of the ontology to plants and animals (at least)---thus also  excluding, among many topics, building architecture---and requires both type-level and instance-level information. 
\item {\em Validation} whether the right content is covered, which includes both classes/ relationships/attributes of the domain and relevant axioms; e.g., querying the African Wildlife Ontology for ``Which plant parts do warthogs eat?'', it requires from the ontology that it contains the entities named {\sf Plant part} (with subclasses), {\sf Warthog}, and either the property {\sf eats} or its reified version {\sf Eating}, or their synonyms, and suitable axioms that relate these entities, such as, e.g., ${\sf Warthog \sqsubseteq \exists eats.PlantPart}$. 
 \item {\em Alignment of a domain entity to a foundational ontology entity}; e.g., and instantiating one of the BFO alignment questions from \cite{BKKM23},  ``Is each warthog wholly present at different times?'', where as `yes' means warthogs are at least a type of endurant. 
\item {\em Elucidation or interrogation of ontological characteristics} of the entity; e.g., ``Is each instance of Tree necessarily (at all times of its existence) an instance of Tree?'' where a `yes' means being a tree is a rigid property of that entity.
\end{compactitem} 
This list of motivations---scoping, validation, alignment of a FO, and property \linebreak interrogation---may not be exhaustive, but already demonstrates not all CQs are alike.

\subsubsection{Components of CQs}

It can be observed from the aforementioned examples that the different CQs have various {\em components}, which are essential for distinguishing between them. 
For instance, the ``unfold in time'' is a description of a key characteristic of perdurants, which are included in a foundational ontology, and there are mentions of specific entities (warthog, Pumba, eats etc.). 
More precisely, elements used by one or more type of CQ can be one or more of:
\begin{compactitem}[--]
\item An existing ontology or a prospective ontology yet to be developed, $o \in \mathcal{O}$, which may be specified as a domain ontology (also including core ontologies and top-domain ontologies), $o_d \in \mathcal{O}_D$, or a foundational or top-level ontology, $o_f \in \mathcal{O}_F$; e.g., the IDO  and BFO, respectively;
\item Domain entity, $d \in \mathcal{D}$, in a subject domain; e.g., SARS-CoV-2 virus, running;
\item Subject domain $ s \in \mathcal{S}$; e.g., building architecture, infectious diseases, data mining;
\item Entity in the ontology, $e \in \mathcal{E}$; e.g., {\sf dolce:Perdurant} in the DOLCE foundational ontology and {\sf awo:lion} in the African Wildlife Ontology;
\item Properties, be they `meta', like being a Sortal, Mixin etc., $n \in \mathcal{N}$, or `metameta', $m \in \mathcal{M}$, such as rigidity and unity that determine whether the entity is sortal etc., or a relational property, $r \in \mathcal{R}$, such as transitivity and symmetry.
\end{compactitem}

Only some of the elements referred to in the last two items are de facto closed sets, or at least may be assumed to change only rarely. Notably, the foundational ontology entities are expected to remain relatively static. The meta properties in $\mathcal{N}$ from \cite{Guarino00} are Type, Quasi Type, Material Role, Phased sortal, Mixin (all Sortals), Category, Formal role, and Attribution (non-Sortals). The main ones in $\mathcal{M}$, or mostly used, are rigidity (rigid, non-rigid, anti-rigid), identity (carrying or supplying an identity criterion), unity, and dependence (externally or not), as underlying the OntoClean method \cite{Guarino00a}, but others are being explored on relevance for ontology engineering, such as telicity (telic or atelic) for processes \cite{Guarino24}. Relational properties in $\mathcal{R}$ include transitivity, irreflexivity, reflexivity, symmetry, asymmetry, and also antisymmetry, acyclicity, and intransitivity beyond the OWL ecosystem  \cite{Keet12ekaw} and possibly also purely reflexive and strongly intransitive \cite{Halpin11}.

\subsection{Descriptions and definitions of the types of CQs}
 
Based on the motivation and component analyses, we introduce an inclusive description for CQs and, being the principal focus,  we identified five main types of CQs for ontologies: those for scoping (SCQ) and validating (VCQ) content of an ontology, to align to a foundational ontology (FCQ) or they concern metaproperties (MpCQ) and those revolving around representing relationships (RCQ). 
We do not exclude the possibility that, in time, more types of CQs for ontologies may be identified. 

\begin{defs}[Competency Question]
A {\em Competency Question} ({\em CQ}) regarding an actual or prospective  {\em ontology} $o \in \mathcal{O}$ is a specialised type of question. The purpose of a {\em CQ} in developing $o \in \mathcal{O}$ may include: scoping, validation or verification, foundational ontology alignment, metaproperty analysis, or relationship assessment or representation. 
\end{defs}

This description of any sort of competency question for ontologies is a generic definition, as it is intended only for setting the context rather than being overly prescriptive at this stage. The specific types of CQs defined below each have distinguishing properties to assist identifying and categorising them, as well as future assistance for authoring questions as to which elements should appear in a question of that type. We provide both a definition of each type and a diagrammatic model snippet in EER notation, where the latter serves the purposes of 1) serving as basis for designing a database or knowledge graph for CQs for analysis, use, and reuse, 2) highlighting the essential components irrespective of what the best language of formalisation would be, and 3) communication for understandability by non-ontologist CQ authors. 
The latter consists of the EER diagram snippet and a controlled natural language rendering of it, possibly augmented with an addition in writing that is strictly beyond ERD notation. Their respective formalisations follow the usual pattern of logic-based reconstructions of EER (see, e.g., \cite{FK24}).

\begin{figure}[t]
  \centering
  \includegraphics[width=0.8\linewidth]{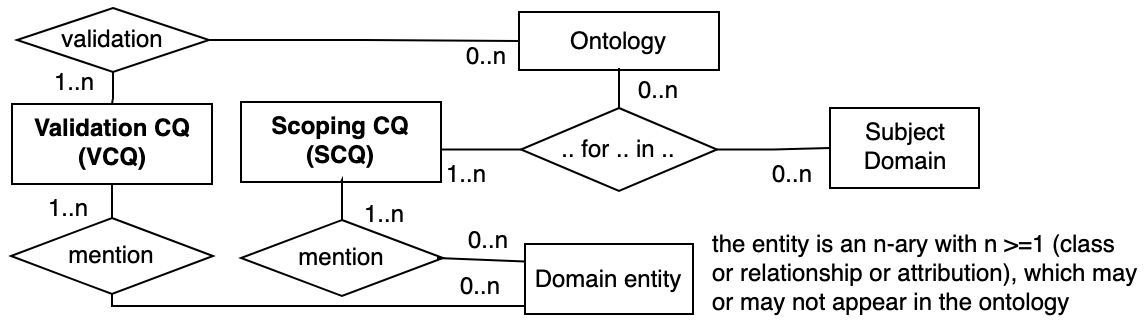}
  \caption{Illustration of the key aspects of SCQs and VCQs for ontologies, in EER notation with annotation.} \label{fig:scq}
\end{figure}

We shall commence with the CQs that ontology developers are familiar with, being the {\em Scoping CQ} (SCQs). They provide a rough demarcation of the possible worlds or interpretations for the ontology or at least the so-called search space for candidate content to be added to the ontology or to be reused from an existing ontology. 
It is graphically sketched in Figure~\ref{fig:scq}.

\begin{defs}[Scoping CQ]
A {\em Scoping CQ} mentions at least one {\em domain entity} $d \in \mathcal{D}$ and contributes to describing the scope for an {\em ontology} $o \in \mathcal{O}$ in a {\em subject domain} $s \in \mathcal{S}$. 
\end{defs}

\noindent Formally, and with the vocabulary as primitives: $\forall x (SCQ(x) \rightarrow \exists y,z,w (mention(x,y) \land DomainEntity(y) \land usage(x,z,w) \land Ontology(z) \land SubjectDomain(w))$, where $Ontology$, $DomainEntity$, and $SubjectDomain$ are to be understood as described in Section~\ref{sec:prelim}, and $mention$ and $usage$ as primitives with their usual natural language meaning that can be suitably formalised accordingly.  
For instance, for SQC for ``Which animals are endangered?'', the $Ontology$ is the African Wildlife Ontology \cite{Keet20awo}, the $DomainEntity$ involves the animals, and $SubjectDomain$ is African Wildlife.  
SCQs typically would be expected to have non-empty answers with the elements asked for. For instance, 
for ``Which animals are endangered?'', the answer set should contain animal species and only those that are endangered. 
They may be answers taken from the TBox or the ABox, as applicable. For an SCQ as a scoping device for an ontology's content, the actual answer is not relevant. It may assist CQ authoring if at least one sample answer is provided, if only to test that the question can be answered.

A {\em Validation CQ} (VCQ) may sound as if it were the same as, or else a kind of a SCQ, or the set of VCQs for an ontology to be a subset of the set of SCQs, since most SCQs are being reused for validation. There are subtle differences, however, both with respect to intent or purpose (validation versus scoping) and they must be answerable by the ontology. SCQs need not be answerable, or even formalisable, whereas VCQs should be both formalisable and answerable. VCQs put testable constraints on the possible worlds or interpretations for the ontology (as logical theory).

\begin{defs}[Validation CQ]
A {\em Validation CQ} mentions at least one {\em domain entity} $d \in \mathcal{D}$ and is used for validating the content of an {\em ontology} $o \in \mathcal{O}$. 
The expressiveness required for formalising the VCQ does not exceed that of the language used to represent the ontology. 
\end{defs}

\noindent The first part of the definition can be formalised as $\forall x (VCQ(x) \rightarrow \exists y,z(mention(x,y) \land DomainEntity(y) \land validate(x,z) \land Ontology(z))$. For the expressiveness constraint, let us consider representation language $L_v$ for the axiom(s) needed to validate the VCQ and $L_o$ either the language used to represent the ontology or the one permitted by the project specification, whichever is more expressive, then $L_v \subseteq L_o$. The reason why this constraint is included is because without it, the VCQ formalisation may be beyond what possibly can be represented in the ontology and thus would never be answerable and thus never be validated, and if it never can be validated, then there is no reason to include that VCQ in the set of CQs for validation.

While SCQs typically will ask for content, and VCQs may as well, VCQs may also be yes/no questions\footnote{at least the VCQs we found do have a number of those type of questions; see ROCQS, introduced in the next section.}, such as any instantiation of mola\_vcq\_6 in ROCQS (adjusted for readability) 
 ``Does (the) Dutch (language) have a region defined (where it is spoken)?'', which is de facto a question about the presence or absence of knowledge in the ontology, not reality, and ``Does OpenOffice meet the ISO-4 standard?'' (instantiation of swo\_41 in ROCQS), which it either does or does not. As can be seen from these examples, they may return content from either the TBox or the ABox, as for SCQs.

By number of CQs that can be found in the literature, the SCQs far outnumber the others, to which we shall return  in Section~\ref{sec:use}. The only other ones we could find for which there were at least two sets of questions that are CQs are those used to align an entity already in a domain ontology to an entity in a foundational ontology, {\em Foundational CQ}s (FCQ). They can be summarised as in Figure~\ref{fig:fcq} and described as:

\begin{figure}[t]
  \centering
  \includegraphics[width=0.9\linewidth]{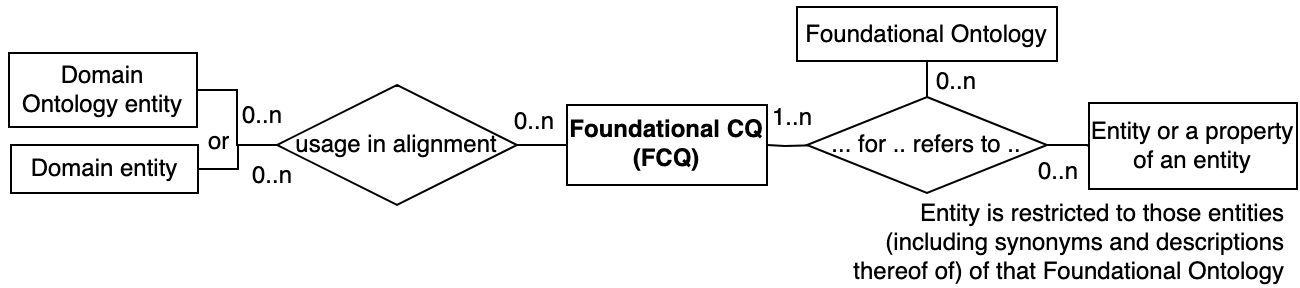}
  \caption{Illustration of the key aspects of FCQs, which are used to interrogate an entity to be aligned to an entity in a foundational ontology.} \label{fig:fcq}
\end{figure}

\begin{defs}[Foundational CQ]
A {\em Foundational CQ} for a {\em foundational ontology} $o_f \in \mathcal{O}_F$ refers to an {\em entity} $e \in \mathcal{E}$ or characteristic thereof that is in the vocabulary of $o_f$. 
A FCQ is intended to be used to in the alignment process of a {\em domain ontology entity} $e^{\prime} \in \mathcal{E}$ (with $e \neq e^{\prime}$) to an entity $e^{\prime\prime} \in \mathcal{E}$ that is in the vocabulary of $o_f$ (where $e$ may be the same as $e^{\prime\prime}$, but need not). An FCQ may also be used to interrogate a domain entity $d \in \mathcal{D}$ independently.
\end{defs}

\noindent The first sentence can be intuitively formalised as (and with abbreviations due to space and readability):  $\forall x (FCQ(x) \rightarrow \exists y,z (referent(x,y,z) \land FoundOnto(y) \land Element(z) \land (containsVocab(y,z) \lor \exists w (containsVocab(y,w) \land hasProperty(w,z)))))$. While we could have chosen to push it into second order, since it refers to properties of elements, we chose not to, because for any implementation it will have to be pushed down to the instance level and would be stored as values in a database. The second part of the definition can be formalised in multiple logically equivalent ways; here, we reify the alignment relationship, resulting in:  
$\forall x (FCQ(x) \rightarrow \forall y,z (purpose(x,y) \land Alignment(y) \land $ \linebreak $ aligningOf(y,z) \land (DomainEntity(z) \lor \exists w (Ontology(w) \land containsVocab(w,z)))))$.

Depending on the controlled natural language that may be designed for the FCQs, they may or may not have a slot for $d$ or $e^{\prime}$ in the sentence. For instance, ``Does it unfold in time?'' can be disambiguated from the context when it is used, but one also can  design a template ``Does [...] unfold in time?'' where the [...] is replaced with the name of the entity that is being aligned ($e^{\prime}$). The `unfold in time' description refers to {\sf perdurant} of the DOLCE ontology, and thus is an example of a property $z$ of entity $e$. 

Answers to FCQ questions may be either yes/no/not-applicable or any other answer option offered, where the consequence is arriving  or staying at  a certain entity in the foundational ontology rather that retrieving answers from the ontology as for SCQs and VCQs.

\begin{figure}[t]
  \centering
  \includegraphics[width=0.95\linewidth]{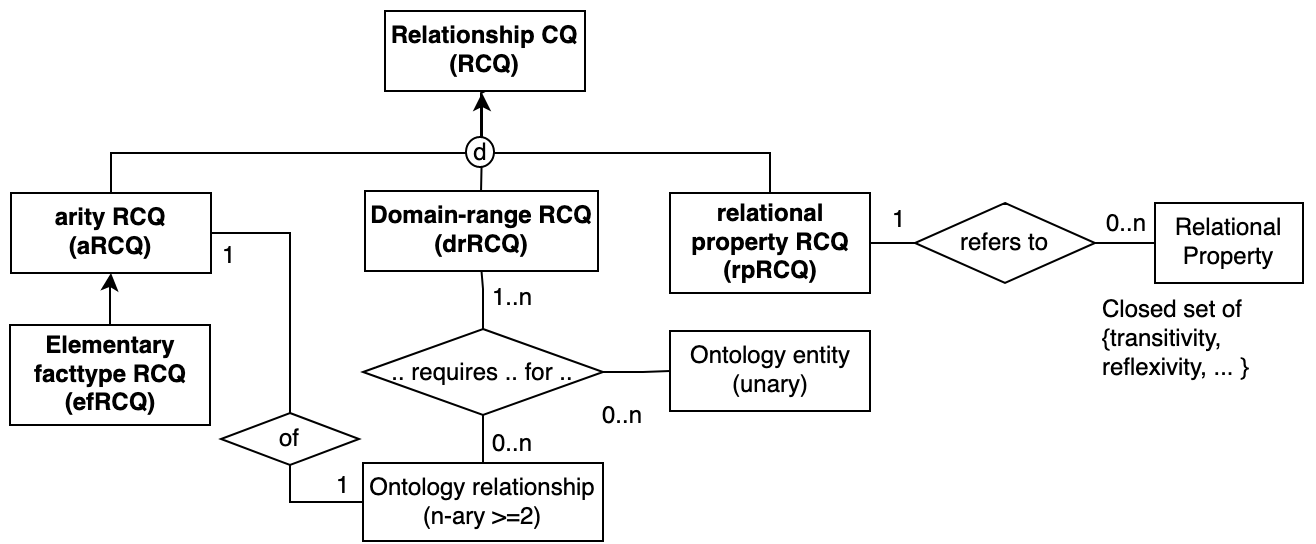}
  \caption{Illustration of the types of RCQs for ontologies and their salient aspects.} \label{fig:rcq}
\end{figure}

The {\em Relationship CQs} are underdeveloped, albeit known informally to some extent. They deal with the arity, or number of participants, of the relationship and ORM's notion of elementary fact type, what the relationship's domain and range (i.e., participants) are, and whether one or more properties, such as transitivity, hold. Since they concern distinct aspects, we created subtypes of RCQs, as indicated in Figure~\ref{fig:rcq} and described in the definition.

\begin{defs}[Relationship CQ]
A {\em Relationship CQ} can be used to determine several key characteristics of a relationship, being:
\begin{compactitem} 
	\item a relationship's {\em arity} (arity CQ, aRCQ), where the relationship has an arity of at least 2, 
	and at least two participants $d, d^{\prime} \in \mathcal{D}$ (which may already be represented in the ontology as $e, e^\prime \in \mathcal{E}$);
	\item whether the relationship is elementary (efRCQ) and not further decomposable without losing information; 
	\item its participating {\em entities} (domain-range CQ, drRCQ), which requires an {\em ontology entity} $e \in \mathcal{E}$ that is a unary (class/concept) for the $n$-ary {\em relationship}
	$e^{\prime} \in \mathcal{E}$ (with $e \neq e^{\prime}$) 
	in the {\em ontology} $o \in \mathcal{O}$; or
	\item its {\em relational property} (relational property CQ, rpRCQ).
The relational property $r \in \mathcal{R}$  asked about in a rpRCQ for a particular ontology $o  \in \mathcal{O}$ should be one expressible in the representation language of $o$.
\end{compactitem}
\end{defs}

\noindent Formally, and also here pushing it into FOL for the same practical reasons as for FCQs, $\forall x (RCQ(x) \rightarrow \exists^{=1} y \exists z (mention(x,y) \land Relationship(y) \land hasParticipant(y,z)))$. For efRCQ, the negation is less easy to capture effectively and the three sample questions are too few and diverse to extract features from. For instance, Halpin's ``Can you rephrase the information in terms of a conjunction?''  \cite{Halpin01} and inspired by an ORM context, ``Does the uniqueness constraint span n or n-1 roles?''. Functional dependencies have a rich literature in database theory, but it plays no role when one can represent only binary relationships like in OWL, and therefore we leave it without additional properties for the time being. The drRCQ can be formalised as follows $\forall x (drRCQ(x) \rightarrow \exists^{=1} y \exists z (requires(x,y,z) \land Relationship(y) \land Class(z) \land Ontology(w) \land$ \linebreak $ containsVocab(w,y) \land containsVocab(w,z)))$. Finally, the rpRQC has only the basic, but unique among other types of CQs, reference to any one element of the closed set of relational properties from $\mathcal{R}$, hence, we obtain $\forall x (rpRQC(x) \rightarrow \exists^{=1} y (referent(x,y) \land RelationalProperty(y)))$.

Unlike the SCQs and VCQs on the one hand and FCQs on the other, as RCQ without the further subtyping, they can be either domain CQs or ontological CQs. Compare, e.g., determining the domain and range of the {\sf employs} relationship, which normally takes domain ontology entities such as {\sf Organisation} and {\sf Worker}, since {\sf employs} is concrete, versus asking about its relational property, such as ``If an organisation employs a worker, must it then never be the case that the worker employs the organisation?'', indicating a broader reach of the relationship. The drRCQs ought to have a universal reach as well, but practically, a number of ontologies have domains and ranges declared such that they are relevant for that particular ontology only.

Expectations of RCQ answers are again different, depending on the type of RCQ. For instance, 
``If a human loves, must it also love itself?'' is expected to be answered by the modeller in the first instance, rather than be obtained from querying the ontology, and the answer is either reflexive or irreflexive, i.e., a property of the relation implied by `loves', but it could be obtained from an ontology to check whether it has been declared. In contrast, an aRCQ like ``how many participants does a loving relation have?'' must have a number as answer, which a SPARQL or SPARQL-OWL or DL query will not be able to answer or is trivially satisfied (all object properties have two participants).

Last, the {\em Metaproperty CQ} (MpCQ) questions refer to a small, probably closed, set of metaproperties about the entities where intentionally the answer will be an absolute that holds across ontologies and, as with the RCQs, they seek to constrain the possible worlds or interpretations. Upon answering, one classifies the entity in a particular type of metaproperty. For instance, if ``Are all persons necessarily always a person during their existence?'' is answered in the affirmative, then that entity, {\sf person}, is a Sortal. A short definition of MpCQs is as follows:

\begin{figure}[t]
  \centering
  \includegraphics[width=0.9\linewidth]{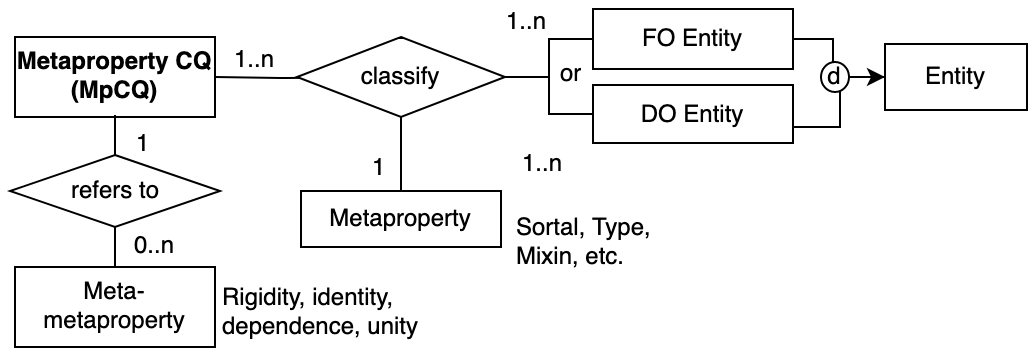}
  \caption{MpCQs for ontologies in EER notation, annotated with typical examples.} \label{fig:mpcq}
\end{figure}

\begin{defs}[Metaproperty CQ]
A {\em Metaproperty CQ} refers to exactly one {\em metameta- property} $m \in \mathcal{M}$  and is used to classify a foundational or domain {\em entity} $e \in \mathcal{E}$ or $d \in \mathcal{D}$
 into, or as being a {\em metaproperty} $n \in \mathcal{N}$.
\end{defs}

\noindent The structure of an MpCQ is sketched in Fig.~\ref{fig:mpcq}. While we expect it to be used for a specific entity in either a foundational or domain ontology, this need not be the case. 
The diagram and definition can be formalised as: $\forall x (MpCQ(x) \rightarrow 
\exists^{=1} y \exists z,w (referent(x,y) \land Metametaproperty(y) \land classify(x,z,w) \land MetaProp(z) \land Entity(w)))$.

MpCQ answers are alike the rpRCQ ones, but then the single element answer is, e.g., `telic' or `anti-rigid', taken from $n \in \mathcal{N}$ rather than $m \in \mathcal{M}$. Considering expressiveness of popular ontology languages, it is expected it will be mostly modellers answering this question rather than querying the ontology for it.

In closing, comparing the scope of each type of CQ, note that SCQs and VCQs apply to an ontology as a whole (however the ontology answers the CQ), and VCQs may be specific to one particular ontology. This is arguably likewise for drRCQs, especially for domain-level relationships. FCQs, if linked to a particular FO, principally have that ontology as scope or relevance, albeit assuming applicability across ontologies. In contrast, the ontological CQs---aRCQ, efRCQ, rpRCQ, MpCQ---relate to an entity and the answer to the CQ ought to necessarily hold across all ontologies, although it may be that some ontology may not have the answer explicitly represented in the logical theory but only in the annotation due to expressivity limitations.

\section{Usage of the Model for Competency Questions}
\label{sec:use}

We will introduce a user story and potential broader use on how the QuO can assist first and describe how the CQs relate to methodologies, touch upon faulty CQs, and finally mention the repository of CQs, ROCQS.

\subsection{Broad intended use: examples}
\label{sec:usecase}

We commence with a user story about ontology development, first introducing some the problems an ontologist may encounter, and then look at where and how the various types of CQs can be of use. 

\begin{quote}
\begin{small}
Sarah, who has a passion for coffee and a growing interest in its complexities, wishes to integrate a coffee ontology in an information system. Sarah wants to design CQs but needs to make considerations as follows. Sarah is uncertain about what information should be included and how it is intended to be used. Ensuring that the ontology accurately represents diverse sensory experiences, flavour profiles, and preferences can be subjective and difficult to validate objectively. She is unclear on whether to draft CQs to integrate her domain ontology with existing foundational ontologies. Formalising complex relationships, such as a transitivity constraints, poses challenges. Analysing and implementing these relationships requires consideration and may lead to unintended implications if not formalised accurately. It is unclear whether there should be CQs around characteristics of coffee, such as rigidity. How could Sarah ensure that a coffee bean from Nicaragua remains as such regardless of processing, blending, or even slight name variations? 
\end{small}
\end{quote}

Using the formal definitions in Section~\ref{sec:model}, Sarah, from the user scenario, generates CQs as follows.
\begin{compactitem}
\item \textbf{SCQs}: Sarah uses SCQs to demarcate the subject domain, with a focus on domain entities about coffee beans and brewing techniques. Examples: 
	\begin{compactitem}[-]
	\item What types of beans are used for coffee? (Arabica, etc.) 
	\item Which equipment is required for each brewing technique? (french press, aeropress, etc.)
	\end{compactitem}
\item \textbf{VCQs}: Sarah uses VCQs to ensure the knowledge contained in the ontology reflects accurate information from the coffee domain using domain entities. Examples: 
	\begin{compactitem}[-]
	\item Is there affogato drink in the knowledge base? 
	\item Which equipment is required for steeping coffee?
	\end{compactitem}
\item \textbf{FQCs}: Sarah uses FCQs to align her domain entities with entities contained in a foundational ontology, ensuring clarity and semantic interoperability. 
Examples: 
	\begin{compactitem}[-]
	\item Can the `brewing method' entity be mapped to any perdurant branch entities in the DOLCE foundational ontology? 
	\item Does `pouring (a drink)' require some further process to be realised? 
		\end{compactitem}
\item \textbf{RCQs}: Sarah employs RCQs to analyse the key characteristics of relationships within the ontology. Examples: 
	\begin{compactitem}[-]
\item Is the `affects taste' relationship between brewing method and coffee strength transitive? (e.g., if method A produces stronger coffee than B, and B stronger than C, does A necessarily produce stronger coffee than C)? (rpRCQ) 
\item How many participants are involved in the relation where a Barista gives a customer a beverage? (aRCQ)
	\end{compactitem}
\item \textbf{MpCQs}: Sarah uses MpCQs to classify entities based on their persistence and existence characteristics. Examples: 
	\begin{compactitem}[-]
\item Is each instance of a coffee bean necessarily (at all times of its existence) an instance of a coffee bean? (rigidity) 
\item Is crema an essential part to the identity of an espresso? (unity)
	\end{compactitem}
\end{compactitem}

\noindent By formulating and answering these diverse CQs, Sarah increases the prospects of developing of a high-quality, accurate, and interoperable ontology for the coffee domain. \\

\begin{figure}[t]
  \centering
  \includegraphics[width=0.90\linewidth]{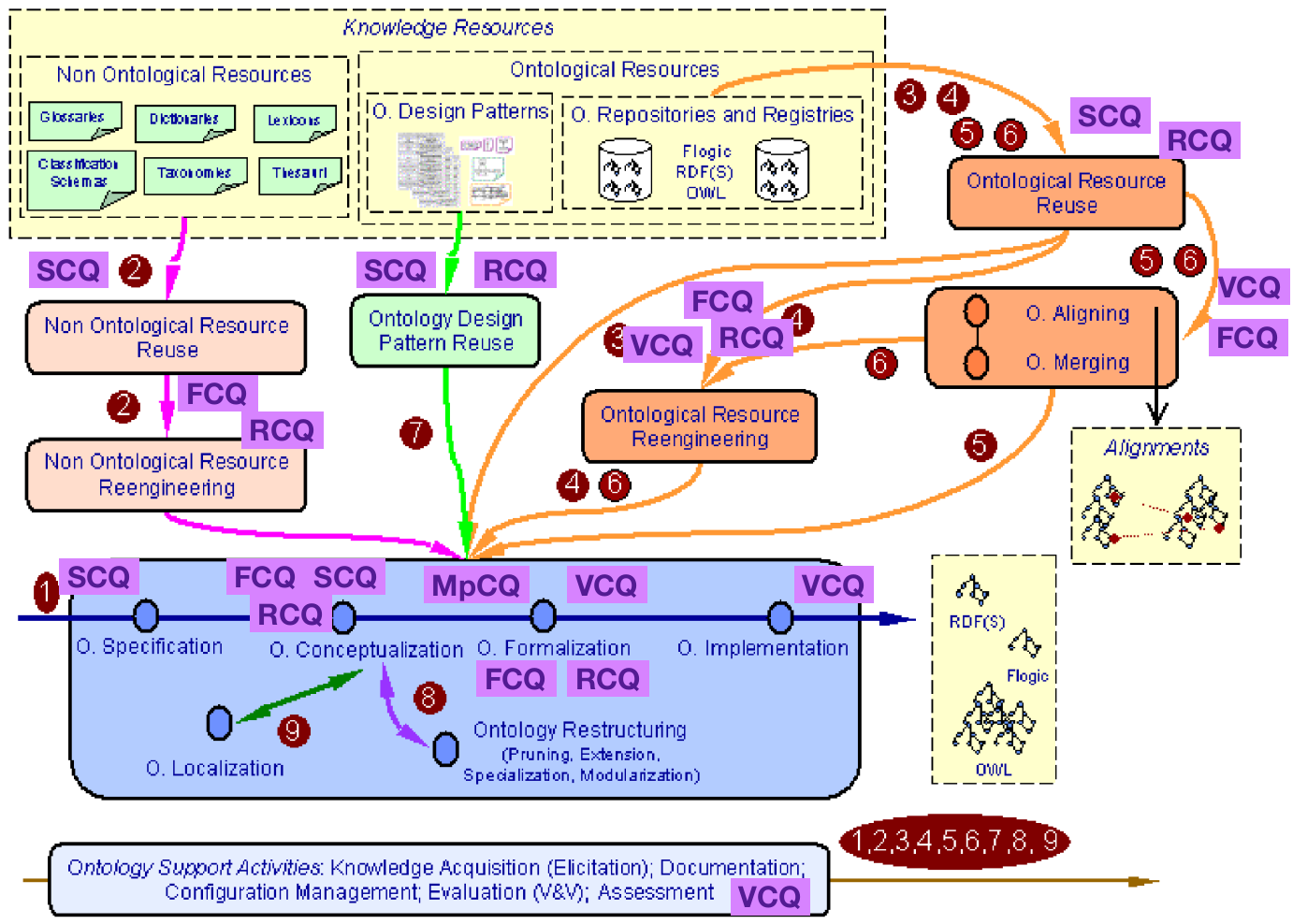}
  \caption{The NeON methodology annotated where diverse types of CQ can be used for various tasks in the process.} \label{fig:neoncq}
\end{figure}

As the user story demonstrates, each CQ type can be relevant for ontology development.
More generally, different types of CQs can be useful at different stages in ontology development.
To illustrate this, we consider the NeON methodology \cite{Suarez08}, among the available options. 
In its summary figure, we indicated {\em where} which type of CQ may be used, as shown in  Figure~\ref{fig:neoncq}, whilst leaving the {\em how} for future work.
Among others, SCQs are used in delineating the subject domain and setting the scope of the ontology (1. specification, 2./3. (non-)ontological resource reuse). As ontology development progresses, VCQs become essential for ensuring the accuracy of information within the ontology (1. implementation). FCQs facilitate the alignment of domain entities with foundational ontologies (5. aligning, 1. formalization, 8. restructuring), enhancing clarity and semantic interoperability. RCQs focus on analysing the key characteristics of relationships within the ontology (1. formalization, 7. ODP reuse). They help formalise relationships in a way that promotes semantic interoperability, addressing issues related to the nature of relationships. MpCQs are relevant to classify entities based on their persistence and existence characteristics (1. formalization).

Besides CQ authoring for a domain ontology, one may consider CQs more broadly, such as possible consequences for an ontology-driven information system. For instance, CQs can link ultimately to, say, an advanced NLP system  to improve the accuracy and relevance of various tasks that are mediated by a linguistic ontology.
For instance, part-of-speech (POS) tagging, which assigns to each word a tag, such as noun, verb, and adjective. What the system is allowed to do, or prevent a POS tagger to make mistakes, may be constrained by the ontology.  In the development of, or assessment whether to use or not, a particular ontology, say, OLIA \cite{chiarcos2015olia}, there may be a MpCQ {\it``Must each annotation of a lexical unit (word) with a POS tag necessarily be NOT annotated with that POS tag for some time of the duration of the existence of that annotation relation?''}. It interrogates about the metaproperty rigidity, with the following elements of the MpCQ definition: {\em metametaproperty} $m$ is Rigidity and domain  {\em entity} $e$  is {\sf Lexical Unit}.  Rigidity refers to whether a lexical unit might change POS tag. If the annotation relation were to be rigid (and tags disjoint), then any word that would be tagged differently across texts (e.g., `bat' tagged as a noun (the animal) and elsewhere as a verb (to hit)) would contradict the knowledge in the ontology. Bat, the animal, is rigid; annotations of lexical units with their part-of-speech are not---a confusion of the two would result in problematic POS tagging software.
Another staple of NLP tasks is morphological analysis of words; e.g., where a word `working' will be analysed and split up into `work' as stem + `ing' as gerund. An ontological assessment of that process, be it for an NLP ontology or the ontology-driven information system derived from it, may involve a CQ {\it ``Is something able to be present or participate in a morphological process?''}. This adheres to FCQ components for the DOLCE foundational ontology referring to its {\sf Event}, with usage in alignment of OLIA's domain ontology entity {\sf morphological process}: an event needs an object, therewith imposing a constraint on a prospective algorithm that it must have objects related to it. 
By propagating CQs and their answers into NLP systems, one can further ensure that they are better aligned with the theories of natural language, which is expected to contribute to better linguistic processing. The argument may analogously be extended to other ontology-driven applications.

\subsection{Exemplary CQs, `almost CQs', faulty CQs, and answerability}

CQs can be problematic or `faulty' i) due to syntax and semantics issues, ii) contextually due to expressivity mismatches of the query language or the ontology language, iii) due to the ontology's content, and iv) relative to the type of CQ. Due to space limitations, we briefly consider only the general idea how they manifest for causes ii and iv.

To illustrate cause iv: FCQs may be faulty in one principal way: it asks to make distinctions that are not made, or: made differently, in the foundational ontology it was designed for and so it is of no use. 
For instance, to ask about process histories when trying to align the domain entity to DOLCE---it does not have {\sf process history}---or to ask about abstract things when trying to align an entity to BFO, which does not contain {\sf abstract}. Likewise, Sarah's CQ on ``further processes'' for `pouring (a drink)' is useful for BFO, but less so for aligning to an entity in DOLCE. 
It may also be the case that some FCQs are better than others because the questions are better understood by the developer \cite{BKKM23}.  
Regarding bad RCQs, and rpRCQ specifically, they fall at least partially in the second option above: many computational ontology languages have very few relational properties or they are highly restricted in their use. 
Examples for other categories exist likewise, although it was difficult to devise clearly faulty MpCQs.  
What may happen rather, is that a metaproperty does not apply to certain categories of entities but that such an entity is interrogated nonetheless, and then if answer options are only, say, `telic' and `atelic', i.e., without an `not applicable' third answer option, a question may not be perceived to be answerable. For instance, the question ``Is a house's endpoint determined outside that process?'' is nonsensical in an endurantist commitment, since {\sf House} is an endurant, not a perdurant and even less so a process. Conversely, Sarah's rigidity question for the coffee bean is relevant in the endurantist commitment.

There are also questions that seek to verify constraints, asking whether content is covered right. These are not  CQs but rather general constraint-checking queries and we list a few types here.
	\begin{compactitem}
	\item Checking for disjointness axioms between classes; e.g., ``Are there animals that are both a carnivore and a herbivore?'', where a `no' means verifying that the ontology has or entails such an axiom or, when used in development mode, to add ${\sf Carnivore \sqcap Herbivore \sqsubseteq \bot}$ if not already present or entailed.
	\item Checking cardinality constraints; e.g., ``Do humans have more than four limbs?'' where upon a `no' it must not be the case that the ontology contains a qualified cardinality constraint ${\sf Human \sqsubseteq \, \geq}\, n\, {\sf hasPart.Limb}$ where $n \geq 4$ without also ${\sf Human \sqsubseteq\, =4 \,hasPart.Limb}$ or ${\sf Human \sqsubseteq \leq 4 \,hasPart.Limb}$.
	\end{compactitem}

\subsection{The Repository of Ontology Competency QuestionS (ROCQS)}

To contribute to further analysis of bad, good, and exemplary questions, we created the Repository of Ontology Competency QuestionS (ROCQS), which consists of 438 CQs covering all types described in this paper. ROCQS conforms to Findable, Accessible, Interoperable, and Reusable (FAIR) principles to enhance their usability in the ontology development process as follows. It is publicly available\footnote{\url{http://www.meteck.org/files/ROCQS/ROCQS.htm}}, the CQs are linked with the specific ontologies they assess for context and understanding, and the coffee use-case is meant to encourage correct usage of CQs. 
In the near future, we plan to enable edit access of ROCQS, either through a Wiki or a portal as a front with a database back-end based on the EER diagram snippets presented in Section~\ref{sec:cqdefs}.

\section{Discussion}
\label{sec:disc}

This first attempt at disentangling CQs for ontologies showed that on the same basis, more types of CQs could be devised, such as the MpCQs, where each of them is plausibly relevant for one or more tasks in the ontology development processes well beyond only scoping and validation. 
While there were ample SCQs and VCQs readily available, perhaps also thanks to introduction and initial framing of CQs for ontologies as SCQs and VCQs in \cite{Uschold96}, we needed to design templates and generate CQs for other categories. Which type will have a greater overall impact on ontology development and quality is yet to be investigated as future work.

The application of CQs in the ontology development process, as demonstrated with Sarah's coffee ontology use case, provides guidance applicable to various domains. The use of ROCQS may also be facilitated by extending it with more questions than the current, to be best of our knowledge, largest collection of CQs. 
ROCQS may also help to keep track of perceived to be easier or harder questions. 
Understandability of CQs was noted in \cite{BKKM23}, in that different wordings or sentence structures of interrogating about the same modelling choices may affect their practical use. 
Further, it may assist in examining Watson's recent exploration of ``worthwhileness'' of the information-seeking act of questions \cite{Watson22} and what worthwhile may mean for CQs for ontologies.

\section{Conclusion}
\label{sec:concl}

The paper presented a model for Competency Questions  within the context of ontology engineering. 
QuO, identifying five main types of CQs, divided into ontological CQs and domain CQs: Scoping (SCQ), Validating (VCQ), Foundational (FCQ), Relationship (RCQ), and Metaproperty (MpCQ) questions. Each type serves distinct purposes within ontology development and they each differ in key constituents of the questions, therewith enhancing the clarity and effectiveness of CQs in ontology development. This, in turn, also assists in determining how CQs can be good or faulty. Diverse usage of the CQs at different stages of ontology development was demonstrated with a use-case and additions to the NeOn methodology. 
To facilitate further research and use of CQs, we created the ROCQS repository with 438 CQs, where the questions are categorised by type, including existing CQ datasets and new sample questions and question templates.

There are multiple avenues for future work, most notably on the effectiveness of FCQs, RCQs, and MpCQs on ontology quality and usability, and prospects for automated generation and evaluation of such CQs, and how to best integrate QuO in the various ontology development methodologies.

%
%
%

%

\end{document}